\documentclass[manuscript,screen]{acmart}
\usepackage{algorithm}
\usepackage{algorithmic}
\DeclareMathOperator*{\argmax}{arg\,max}

\usepackage{graphicx}
\usepackage{subcaption}

\AtBeginDocument{%
  \providecommand\BibTeX{{%
    \normalfont B\kern-0.5em{\scshape i\kern-0.25em b}\kern-0.8em\TeX}}}

\begin{document}

\title{Spatio-temporal Incentives Optimization for Ride-hailing Services with Offline Deep Reinforcement Learning}


\author{Yanqiu Wu}
\affiliation{%
  \institution{New York University}
  \country{USA}
}
\email{yanqiu.wu@nyu.edu}

\author{Qingyang Li}
\affiliation{%
  \institution{DiDi}
  \country{China}
}
\email{qingyangli@didiglobal.com}

\author{Zhiwei Qin}
\affiliation{%
  \institution{DiDi}
  \country{China}
}
\email{qinzhiwei@didiglobal.com}


\begin{abstract}
A fundamental question in any peer-to-peer ride-sharing system is how to, both effectively and efficiently, meet the request of passengers to balance the supply and demand in real time. On the passenger side, traditional approaches focus on pricing strategies by increasing the probability of users' call to adjust the distribution of demand. However, previous methods do not take into account the impact of changes in strategy on future supply and demand changes, which means drivers are repositioned to different destinations due to passengers' calls, which will affect the driver's income for a period of time in the future. Motivated by this observation, we make an attempt to optimize the  distribution of demand to handle this problem by learning the long-term spatio-temporal values as a guideline for pricing strategy. In this study, we propose an offline deep reinforcement learning based method focusing on the demand side to improve the utilization of transportation resources and customer satisfaction. We adopt a spatio-temporal learning method to learn the value of different time and location, then incentivize the ride requests of passengers to adjust the distribution of demand to balance the supply and demand in the system. In particular, we model the problem as a Markov Decision Process (MDP). The problem is solved in two steps: 1) based on historical trip data from the ride-hailing platform, we propose a deep reinforcement learning based method with constrained actions to summarize demand and supply patterns into a Spatio-Temporal network, 2) to solve the budget constraint problem in the spatio-temporal incentive, we formulate an integer programming step in real-time policy learning process, where each state-action pair is valued in terms of immediate reward, future gains and discount budget. Through experiments, we demonstrate that our proposed approach can deliver improvement on the marketplace efficiency.
\end{abstract}




\maketitle

\section{Introduction}
With rapid development of the Global Position System(GPS), and on-demand ride-hailing services such as Uber, Lyft and Didi Chuxing, significant improvements has been achieved over traditional taxi systems. Massive amount of trip data become available, offering much more opportunities for providing more intelligent and convenient services and a surge in passion in research fields such as driving routing planning \cite{route1}\cite{route2}, order dispatching \cite{Glaschenko2009MultiAgentRT}\cite{drlonline}\cite{li2019efficient}, supply chain management\cite{supplychain}, driver program \cite{shang2019environment}\cite{shang2021partially} and demand prediction \cite{yao2018deep}\cite{bikelane}. 

One of the most important task in online ride-hailing platforms is to balance supply and demand situation in advance. 
Once passenger's request has been answered, the trip would bring both the passenger and the driver to the destination, which can be viewed as a process of relocating the driver to the destination. The driver can answer future requests at the new location. For example, when a ride order travels to a hot (high demand) zone where drivers are more needed, after fulfilling the ride, the driver is more likely to receive new requests there. If the situation of traversing to a hotter zone persists, the driver can receive more gain in the long turn. Hence, we see that the value of a trip goes beyond the trip fare and depends also on the spatio-temporal long-term values. 

Dynamic pricing and incentives are one of the levers in the ride-hailing system that can influence both supply and demand distributions to make them more aligned, and hence, improving customer experience through higher driver income and request answer rate. 
Dynamic pricing and spatial pricing are mainly explored by previous researchers. Some work focused on approximate dynamic programming(ADP) \cite{YANG2020126} and model predictive control(MPC) \cite{NOURINEJAD2020340}. Other researchers build up spatial equilibrium \cite{Bimpikis2017, ZHA201858, HE201593}. These research studies have some limitations. For example, they can examine dynamic and spatial pricing separately but not simultaneously. Recently, to achieve spatio-temporal incentivization, some work of reinforcement learning enhanced agent-based modeling and simulation system have been explored \cite{CHEN2021103272}. However, to optimize the reinforcement learning policy, the method requires to build up a data-driven simulation environment to mimic a real-world ride-sourcing platform's operations on a city network to enable online training. 

Recently, offline reinforcement learning \cite{fujimoto2019off}\cite{levine2020offline}\cite{kumar2020conservative} has drawn a lot of attention since it promises to learn effective policies from previously-collected, static datasets without further interaction. Because offline RL does not rely on the built environment, it has achieved great success in many real-world application scenarios, such as recommendation system, real-time decision-making system, advertising optimization, robotics and so on. BCQ \cite{fujimoto2019offpolicy} proposed a batch-constrained reinforcement
learning method close to on-policy with respect to the available data to solve the extrapolation error problem. BEAR \cite{kumar2019stabilizing} is a practical solution to reduce the bootstrapping error accumulation.

However, our problem can not be solved directly with the existing offline RL method. In the real ride-hailing scenarios, when we model the  spatio-temporal incentives optimization problem as a Markov Decision Process(MDP), incentive is treated as the action and it will bring the related cost when optimizing the policy. There are some related work \cite{wu2015algorithms} to study the MDP problem under budget constraints while none of them will work in the offline RL scenarios. It brings more challenges when we constrain the budget of the whole city from a global view, the solution space of this problem will be huge and it becomes a NP-hard problem.

Our work adopts reinforcement learning methods, but instead of training a simulated ride-sourcing platform and an on-policy agent, we formulate our problem in the offline reinforcement learning setting, and utilize previously collected data, without additional online interactions with the environment. The goal for spatio-temporal incentive is not only to fulfill the ride requests of the current passengers, but also to optimize the anticipated future gains. In this study, we propose an offline deep reinforcement learning to learn the spatio-temporal value network to model the demand and mobility patterns for the whole city. In addition, to solve the budget constraint problem during the policy learning, we formulate it as an integer programming process to conduct the policy optimization within the limitation of budget threshold and learn the state-action network simultaneously. To verify the effectiveness of the proposed method, we apply it to a real ride-hailing system. Through comparative evaluations, our proposed method shows significant improvements in the real-world application.

\begin{figure}[t]
\centering
\includegraphics[width=0.50\textwidth]{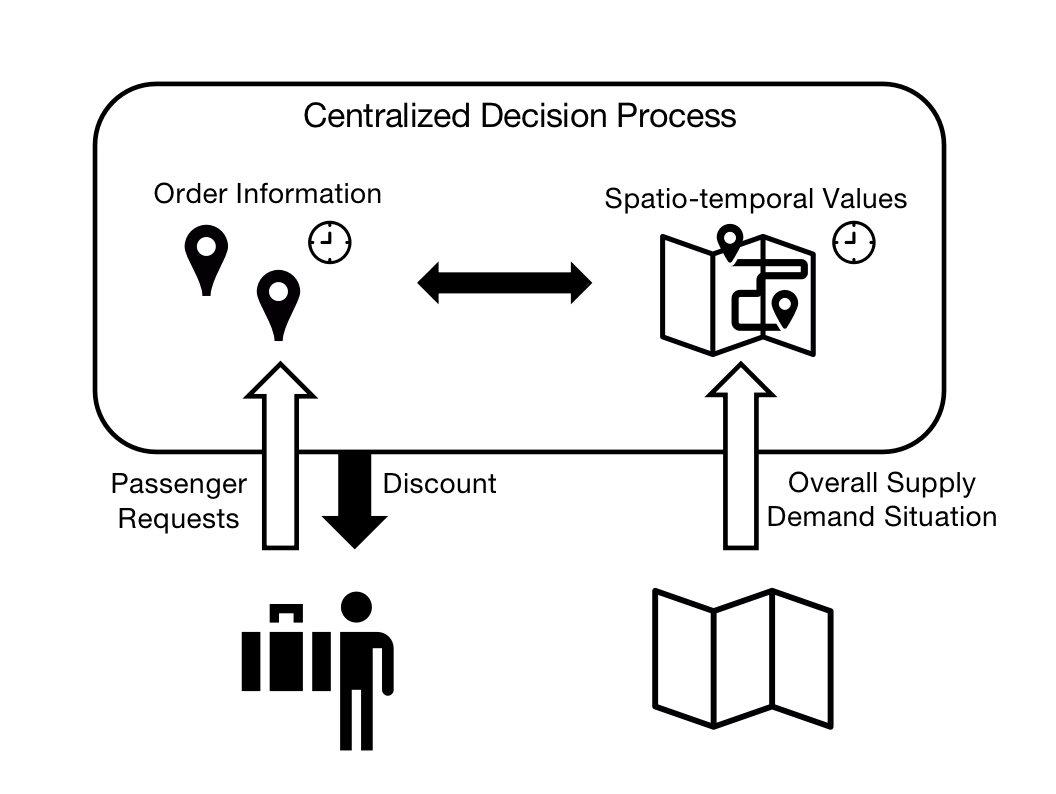}
\caption{Overview of spatio-temporal incentives optimization in ride-hailing services.}
\label{fig:overview}
\end{figure}

The contribution of this work can be summarized as follows:
\begin{itemize}
    \item We propose a novel offline reinforcement learning to dynamically adjust the spatial-temporal distribution of demand to balance driver supply and customer demands. To the best of our knowledge, this is the first work of offline RL to optimize spatio-temporal problem on the demand side. 
    \item  We tackle the budget constraint problem in the offline RL setting by formulating the constraint RL problem as an integer programming process to learn the spatio-temporal value network as well as control the resource consumption of policy.
    \item Experiments are conducted under different times of the week(weekdays, weekends) and from multiple aspects to analyze supply and demand, including comparing to SOTA offline RL algorithms. We largely improve both revenue and efficiency in balancing supply and demand across the city.
\end{itemize}

In what follows, we will describe the passenger pricing problem in section \ref{sec:prob} and formulate our MDP in section \ref{sec:mdp}. The detailed explanation of our deep Q-network approach combined with Integer Programming can be found in section \ref{sec:alg}. The learning and optimization capabilities of our deep reinforcement learning algorithm and the advantage of the integer programming method through a set of experiments in section \ref{sec:exp} using real data from a major ride-hailing platform. We close the paper with a few concluding remarks in section \ref{sec:con}.

\section{Related work}\label{sec:prob}

\subsection{Background and System Overview of Spatio-temporal Incentive}
Recently, on-demand online ride-hailing platform like Uber, Lyft, and DiDi provide a more efficient way of transportation. On the passengers' side, before they actually request a ride order, they need to provide information of the ride, including the starting location, destination and the time of the ride order. We name this stage as ride inquiry. Then, these online ride services have a centralized system to give some information back to the passengers, such as estimated travel time and estimated ride fee. Finally, based on these information, passengers can decide whether they actually want to place the ride order. We name this stage as ride request. The conversion probability from ride inquiry to ride request is defined as estimated-call-rate (ECR). Naturally, while other information of the ride keeps unchanged, the cheaper the fee is, the higher the ecr would be. Once the passenger requests for a ride, the central dispatching system will try to match a driver to pick up the passenger to finish the ride. The probability of a ride being successfully completed by a driver is called completion-rate(CR). CR will not be affected by ride pricing. 

The optimization of passenger pricing and incentive is crucial to balance the demand and supply since the incentive strategy will directly have influences on the distribution of demand. One special case is that when the amount of supply is much larger than the demand, which means there are more idle drivers in this region, the platform should give more discount to passengers to increase the demand amount, helping drivers complete more orders and increase drivers' income. Another case is that when amount of supply is less than the demand, which means there are a lot of trip request in this region. However, different order requests have different destinations, which will take the driver to a different geographic location in the next moment and affect the driver’s future income. If the driver goes from the hot zone to the cold zone, it will reduce the driver’s income from the next order, otherwise it will increase the driver’s probability of taking orders and increase income. Therefore, the optimization of the pricing strategy is crucial to the driver's income, the balance of supply and demand, and the improvement of the overall trading market efficiency.

The incentives system is responsible for distributing real-time discounts to passengers with open trip inquiries. Figure \ref{fig:overview} shows a system overview of how it works in practice.
Every time a passenger enters the ride inquiry stage, a centralized decision system would decide on the discount rate of the current ride. Initially, the decision only affects passenger's ecr of the order. Once the passenger enters the ride request stage and is successfully picked up by a driver, the discount would only takes effect at the payment stage. 

Given both historical ride data and real-time order information, we can potentially optimize the pricing system’s global efficiency in a long horizon by spatio-temporal pricing. When a passenger requests a ride and the order is finished, he or she has moved from the starting location to the destination. Meanwhile, the driver who carries out the ride, is also relocated to the destination. Hence, a trip changes the spatio-temporal state of a driver, which together with demand distributions, forms the basis for future order matching and vehicle repositions. The spatio-temporal state values are the quantities that characterize these long-term effects and can thus guide the incentives decisions to optimize global metrics, e.g., total driver income. 

\section{Learning}
\label{sec:mdp}
The learning step aims to provide a quantitative measure of the spatio-temporal patterns of the ride supply and passenger demand across the whole city. Given historical off-line ride data, we build a Markov Decision Process (MDP) representing rides happening on the platform. The learned value functions provide quantitative values for each spatio-temporal state, which is later used in the discount planning step. 
\subsection{Problem Statement}
\emph{MDP definition.} The MDP we build here is in a global view. In \cite{orderdispatch}, it proposed a tabular based reinforcement learning method to solve the problem of order dispatch in the ride-hailing platform from the perspective of local view, which aims to match drivers with different ride orders, each driver is modeled as an agent. However, in this study, unlike drivers, if a passenger requests a ride from location A to location B, there's no guarantee that he or she will request the next ride starting from location B. Hence, we can not model each independent passenger as an agent. Instead, we adopt the global-view setting which we do not distinguish individual passengers. Each passenger is regarded as the same. For example, if one ride transits from A at time t and arrives at B at time t+1, any ride leaving B at time t+1 can be regarded as our possible next transit.  

When a passenger requests a ride, we have the information of the starting location, destination, time of the request, estimated travel time and estimated ride cost. Our MDP formulation uses these available information. A trip transition consists of order request, pick-up and completion: The passenger enters the origin and destination location. Ride-hailing platforms provide estimated travel time and fee. The trip moves the driver and passenger to the destination. The driver earns an immediate reward (trip fee) from this transition. We list the key elements of our MDP formulation below.

\emph{State.} The state is defined to indicate geo-location information of the starting location and the time of the ride request. For simplicity, we quantize time periods into semi-hours. Formally, we define $s = (g, t, f) \in S$, where $g \in G$ represents the grid region where the trip starts, $t \in T$ is the time index. We also differentiate the time for weekday and weekend. And $f \in F$ contains the additional contextual features, such as the statistics of supply and demand of the current grid $g$. 

\emph{Action.} The action is defined as the discount assigned to the trip. Action space is discrete: 0.75, 0.8, 0.85, 0.9, 0.95, 1, corresponding to 25\% off, 20\% off, 15\% off, 10\% off, 5\% off and no discount of the ride fee. Naturally, as the discount goes up of the ride fee, the probability of a ride inquiry becoming a actual ride request would increase.   

\begin{figure}[t]
\centering
\includegraphics[width=0.50\textwidth]{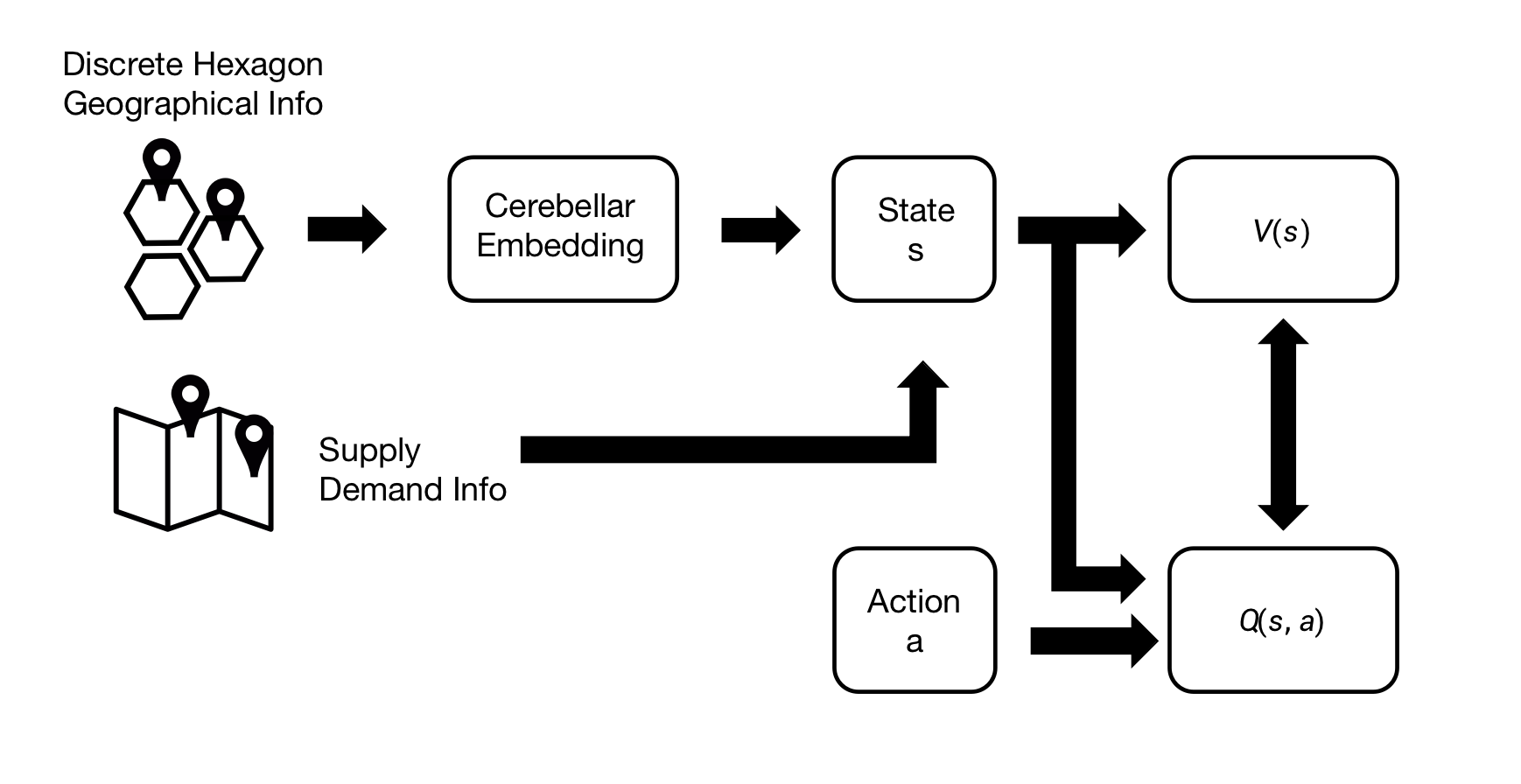}
\caption{Architecture of the proposed algorithm}
\label{fig:input}
\end{figure}

\emph{Reward.} In order dispatch setting, reward is defined as the total fee collected for the trip. Intuitively, in our problem, the reward can be defined as: $r(s_t,a_t) = \delta(ecr_t) * GMV_t$, where $\delta(ecr_t)$ represents the change of ecr\footnote{ECR represents the probability of a user calling an order, which ranges from 0 to 1.} under discount $a_t$ and $GMV_t$ represents the trip fare of trip $s_t$. The optimal value network of a state s, $V^*(s)$, would be the optimal estimated future delta Gross Merchandise Volume(GMV).  

\emph{Episode.} An episode is one complete day, from 0:00 am to 23:59pm. Trips crosses midnight are regarded as terminal state.

\emph{State transition} and \emph{reward distribution.} 
The discount distribution have to be made in a rolling horizon manner. In real work application, at the time of a ride request, the actual trip fare and travel time are unknown. Our system have to depends on the estimated arrival time and estimated price to make decisions. The error distribution of these estimations are implicitly captured by state transitions and reward distributions.  

\emph{Discount factor.} The discount factor in the MDP definition controls how far we look in to the future. In the order dispatch problem, the rewards defined in \cite{orderdispatch} are also discounted. Here in our passenger discount application, we adopt the same discounting methods. For example, for an order which takes $T$ time slots(semi hours) with trip fare $GMV$ and a discount factor $\gamma$, the final reward is given by: 
\begin{equation}
\label{eqn:rew}
    R_\gamma = \delta(ecr) * \big(\sum_{t=0}^{T-1} \gamma^t \frac{R}{T}\big)
\end{equation}
For example, suppose a passenger requests a order from A to B at time 00:00. The trip is estimated to be completed in 40 minutes and costs 40 dollars. Suppose we assigns 10\% off discount to the order, which increase the passenger's ecr by 15\%. We use a discount factor of $\gamma = 0.9$ and segment the time slots into 10-minutes windows. In our model, this order will transit from $(A, 0, f_a)$ to $(B, 4, f_b)$ with action $a = 0.9$ and $r = \delta(ecr) * (10  + 10 * 0.9 + 10 * 0.9^2 + 10 * 0.9^3) = 0.15 * 34.39 = 5.1585$.

\section{Algorithm}
\label{sec:alg}
In this section, we proposed an offline deep reinforcement learning method to learn the long-term spatio-temporal value for the whole city, to dynamically change the distribution of demand and achieve a state of balance between supply and demand, improving the efficiency of the marketplace. For more detailed information of our proposed method, please refer to the pseudocode provided in Algorithm \ref{alg:pseudo} and the architecture overview in figure \ref{fig:input}.

\subsection{Deep Q-network}
To solve the MDP formulation, we adopt the model-free approach and use the deep reinforcement learning framework proposed in \cite{mnih2015humanlevel}. In the deep Q-network framework, the mini-batch update is a step for solving a bootstrapped regression model with the Mean Squared Error loss function:
\begin{equation}\label{eq:qtarget1}
    \Big(Q(s_t, a_t | \theta) - \big(r(s_t,a_t)+\gamma \argmax_{a_{t+1} \in A} Q(s_{t+1}, a_{t+1}| \theta^{'})\big)\Big)
\end{equation}
where $\theta^{'}$ is the weights for the target Q network, $A$ is the action space. 
To improve training stability and avoid overestimation, we use Double Q learning \cite{vanhasselt2015deep}. Specifically, in addition to the Q network, we use a corresponding Q target network. The targets in equation (\ref{eq:qtarget1}) is modified so that the argmax is evaluated by Q network, and the highest valued action is used by the target Q network as follows:
\begin{equation}
    r(s_t,a_t) + \gamma \mathcal{Q}_{targ}(s_t, \argmax_a \mathcal{Q}(s_t, a))
\end{equation}

\subsection{Model Training}
Building a realistic simulator to imitate city-scale driver and passenger dynamics is very challenging. However, a large number of historical trips are usually available. Hence, we use historical trip data for training, which means our training is off-line and we do not interact with any environment during training. The historical data is generated by some existing behavior policy. Each trip $x$ defines a transition $(s_t,a_t,r_t,s_{t+1})$. These data are retrieved from data warehouse and store in a replay buffer for training similar to \cite{drlonline}. During training, each mini-batch is sampled from the replay buffer to train the Q and V spatio-temporal network. 

\subsection{Cerebellar embedding}
Having a good state representation is usually the key step to
solving a practical problem with neural networks. In our spatio-temporal incentives optimization problem, we require the parse of complicated state information as the basis for long term reasoning. Hence, we adopt the cerebellar embedding scheme used in order dispatching problem\cite{tang2021deep}. Cerebellar embedding combines CMACs with embedding to obtain a distributed state representation\cite{tang2021deep} that is generalizable, extensible and robust. CMACs involves multiple overlapping tilings of the state space. The total number of tiles is the size of the conceptual memory. The mapping
from input to tiles is done such that input points close together
in the input space have considerable overlap between their set of
activated tiles. The output of CMACs is computed as the sum
of the weights of the activated tiles.

\begin{algorithm}[t]
    \caption{Spatio-Temporal Network}
    \label{alg:pseudo}
    \begin{algorithmic}
        \STATE Load historical data into the replay buffer $\mathcal{M}$, state $s = (l, t, f)$
        \STATE Ranomly initialize the spatio-temproal network $\mathcal{V}$ with weights $\phi$ and state-action value network $\mathcal{Q}$ with weights $\theta$
        \STATE Initialize the target network $\hat{\mathcal{V}}$ and $\hat{\mathcal{Q}}$ with the same weights as $\mathcal{V}$ and $\mathcal{Q}$
        \FOR {t = 1,2, \dots, T}
           \STATE Sample a random mini-batch $\mathcal{B} = \{(s_j,a_j,r_j,s_{j+1})\}$ from $\mathcal{M}$.
            \STATE Calculate $\mathcal{V}$ target $v_j$ with
            \STATE $v_j = \hat{\mathcal{Q}_{\theta'}} (s_{j}, \argmax_a \mathcal{Q}_{\theta}(s_{j}, a))$
            \STATE Calculate $\mathcal{Q}$ target $y_j$ with
            \STATE $y_j = r_j + \gamma  (1-d) \hat{\mathcal{V}_{\phi'}}(s_{j+1}) $ \COMMENT{d is the terminal state signal}
            \STATE Update $\mathcal{Q}$ by one step of gradient descent on 
            \STATE $\nabla_\theta \frac{1}{|\mathcal{B}|} \sum (\mathcal{Q}_\theta(s_j, a_j) - y_j)^2$
            \STATE Update $\mathcal{V}$ by one step of gradient descent on 
            \STATE $\nabla_\phi \frac{1}{|\mathcal{B}|} \sum (\mathcal{V}_\phi(s_j) - v_j)^2$
            \STATE Calculate advantage matrix $\mathcal{A}$ as in equation \ref{eqn:valm}
            \STATE Calculate cost matrix $\mathcal{C}$ as in equation \ref{eqn:costm}
            \STATE Solve policy distribution $X$ using $A$ and $C$ as in equation \ref{eqn:ip}.
            \STATE Add $\mathcal{B'} = \{(s_j,a_j',r_j',s'_{j+1})\}$ in $\mathcal{M}$ where $a_j'$ are given by $X$.
            \IF {t mod update frequency = 0}
                \STATE Update the target network with:
                \STATE $\theta' \gets \theta$
                \STATE $\phi' \gets \phi$
            \ENDIF
        \ENDFOR
        \STATE return $\hat{\mathcal{V}}$ and $\hat{\mathcal{Q}}$
    \end{algorithmic}
\end{algorithm}

The cerebellar embedding extends CMACs. An randomly initialized embedding
matrix is used as the actual memory and the mapping is implemented
using a sparse representation. It defines multiple tiling functions and each function would map the continuous input to a unique string id indicating one discretized region of the state space. The full details of cerebellar embedding can be found in \cite{tang2021deep}. 

To quantize the geographical space, we also adopt the hierarchical coarse-coding in the location space as in \cite{tang2021deep}. Hexagon is used as the tile shape because hexagons have only one distance between a hexagon center-point and its neighbors. 

\begin{figure*}[h!tb]
\centering
\begin{subfigure}{0.4\textwidth}
	\centering
	\includegraphics[width=0.95\linewidth]{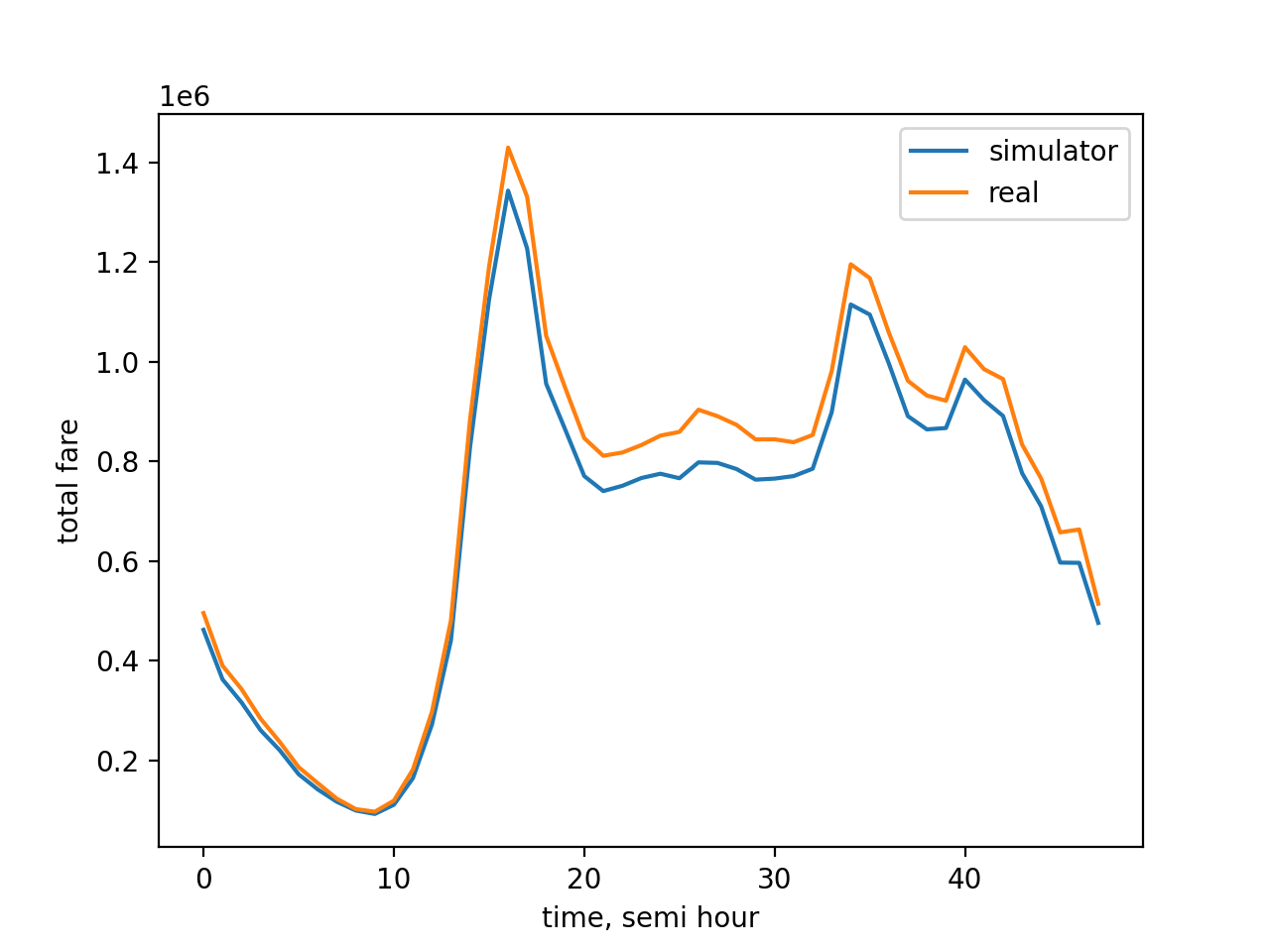}
	\caption{GMV comparison in terms of calling orders}
	\label{fig:ecrsim}
\end{subfigure}
\begin{subfigure}{0.4\textwidth}
	\centering
	\includegraphics[width=0.95\linewidth]{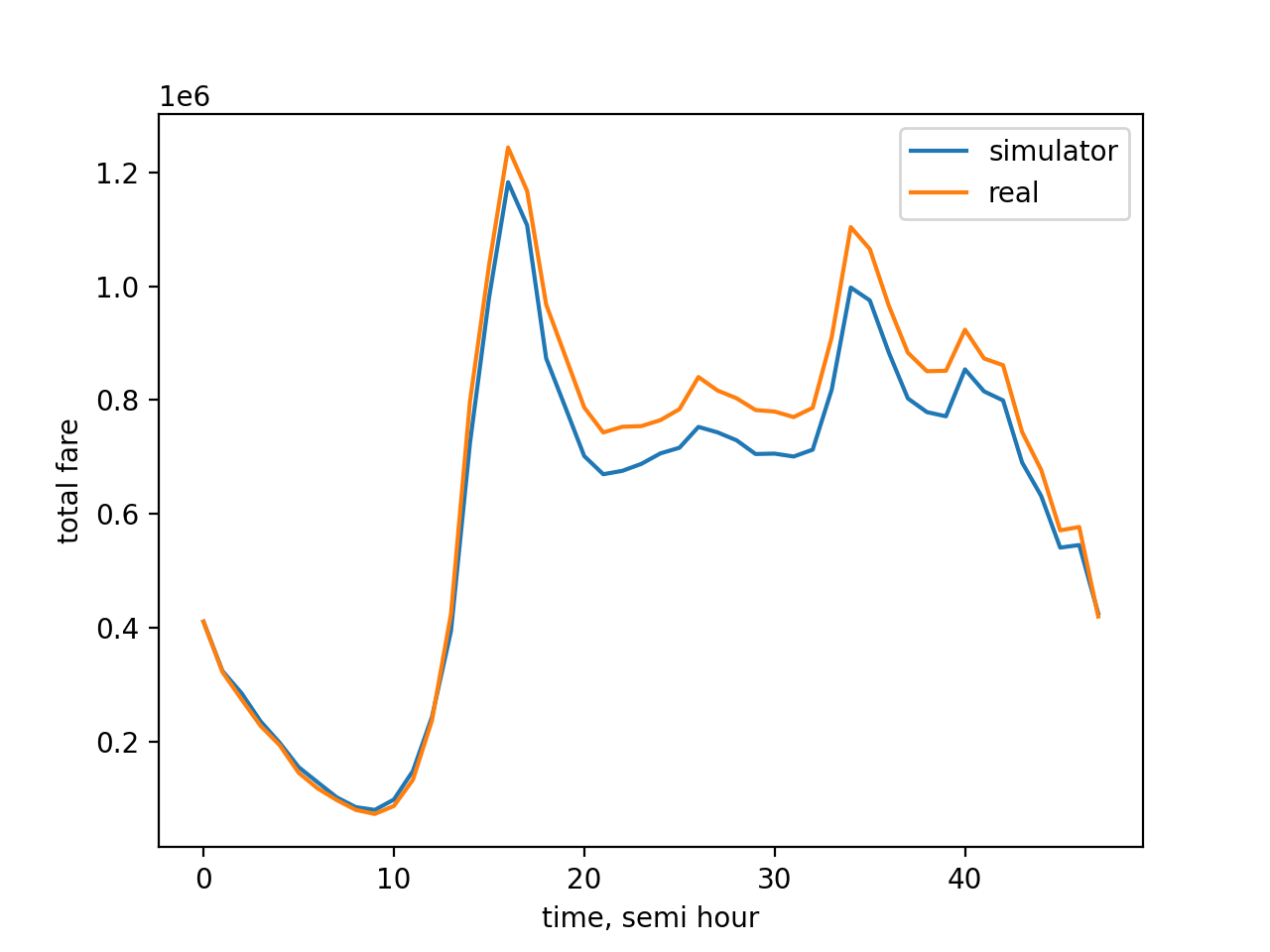}
	\caption{GMV comparison in terms of finished orders}
    \label{fig:crsim}
\end{subfigure}
\caption{The simulator calibration in terms of GMV. The red
curves plot the GMV values of real data averaged in 24 hours of one day, in 30-minute time granularity. The blue curves are simulated results averaged over 48 semi-hours.}
\label{fig:ecr_cr_sim}
\end{figure*}

\subsection{Extrapolation error}
Off-line reinforcement learning has the problem of extrapolation error \cite{kumar2019stabilizing}. Exploration error happens due to mismatch in the distribution of data produced by the policy and the distribution of data contained in the batch. \citet{fujimoto2019offpolicy} introduce batch-constrained reinforcement learning, where agents are trained to maximize accumulated rewards and minimize the distance of selected actions to the data in the batch. Recall that our action space is discrete. Hence, to tackle extrapolation error, we perform a simple action search combined with Q networks, so that selected highest valued action has been seen in the batch. $\hat{A}(s) = \{a | (g(s),a) \in B\}$ where $g(s)$ is the discretized spatio-temporal grid that state s falls into and $B$ is replay buffer. If the action search returns an empty action set, we would randomly select an action instead of the highest valued action to ensure exploration. 

\subsection{Terminal state}
Naturally, we would define the trips across the day as our terminal states. However, in reality, especially in big cities, there are still a large amount of trips happening around midnight. Hence, in order to be consistent with real world situations, we define trips crossing 3:00am as our terminal states. 

\subsection{Integer Programming}
\label{alg:ip}
Recall that our action space is discrete, given the deep Q-network framework, a natural policy is to select actions that maximize the learned Q values. However, in this specific application environment, ride-hailing platforms have a total budget for distributing discounts to passengers. Hence, we need to make sure that discount distribution induced by our policy will not exceed the total budget. Simply selecting the highest valued action does not guarantee this constrain. Instead, we form the optimal discount distribution problem as an Integer Programming problem:
\begin{equation}
\begin{split}
\label{eqn:ip}
    &\text{maximize } A \circ X \\
    &\text{subject to } C \circ X \leqslant \text{B}, X \geqslant 0
\end{split}
\end{equation}
where $A$ represents the value matrix of the trips we want to solve, $C$ represents the corresponding cost for discounts, $B$ is the total budget. 
   
Formally, matrix $X$ is a binary matrix such that for each trip, only one of the 6 actions has the corresponding value of 1. We defined our value matrix $A$ as a N by 6 matrix to represent the different values each trip can get under different actions. N represents the number of trips. $\beta$ is a hyper-parameter used to weigh between the trip fare and the Spatio-Temporal value transitions traveling from the starting location to the destination. cr\footnote{cr denotes the probability that a calling order is accepted by the driver and the order is finally completed. Its value ranges from 0 to 1.} indicates the probability of successful completion of the order.
\begin{equation} 
\label{eqn:valm}
    A =\delta \text{ecr} * \text{cr} * (\beta \text{GMV} + (1-\beta)\gamma (1-d) \mathcal{V}(s') - \mathcal{V}(s))
\end{equation}

The cost matrix $C$ is similarly defined as follows:
\begin{equation}
\label{eqn:costm}
    C = (1 - a) * \text{GMV} 
\end{equation}
where $a$ represents the discrete actions which take values of 0.75, 0.8, 0.85, 0.9, 0.95 and 1. 

By solving the integer programming problem, we receive a discount distribution that optimize for the combination of trip fares and Spatio-Temporal value transitions, meanwhile, we do not exceed budget. 

\section{Experiment}
\label{sec:exp}
In this section, we will discuss the experiment settings and results. We use historical ExpressCar trip data obtained from a major ride-hailing platform as our training data. The data set consists of over three millions of trips happening from April 13th to April 26th, 2020 and is divided into training set (one week) and testing set (one week). Data points in each mini-batch are viewed as samples on the Q value function. We used a discount factor $\gamma = 0.9$ and normalized all state vectors with their population mean and standard deviation. We found the pre-processing was necessary for a stable training. 

To provide a more complete and direct understanding of the effectiveness of the proposed algorithm, we evaluate it from three different aspects, including a environment simulation, algorithm performance comparison and map visualization.

\subsection{Environment Simulation}

\begin{figure*}[h!tb]
\centering
\begin{subfigure}{0.3\textwidth}
	\centering
	\includegraphics[width=0.95\linewidth]{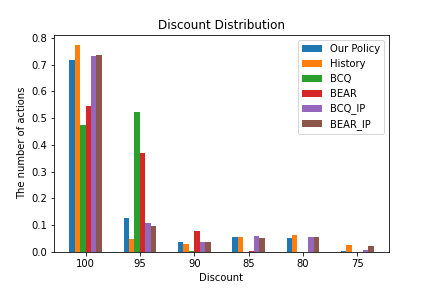}
	\caption{Overall discount distribution}
	\label{fig:all_dist}
\end{subfigure}
\begin{subfigure}{0.3\textwidth}
	\centering
	\includegraphics[width=0.95\linewidth]{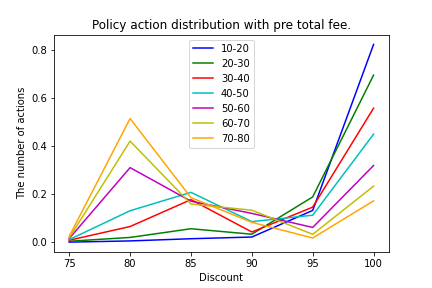}
	\caption{Our policy discount distribution with total fare}
    \label{fig:policy_fare_dist}
\end{subfigure}
\begin{subfigure}{0.3\textwidth}
	\centering
	\includegraphics[width=0.95\linewidth]{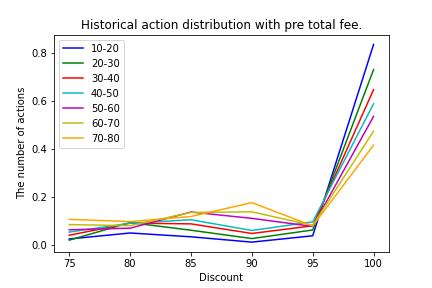}
	\caption{Historical discount distribution with total fare}
	\label{fig:hist_fare_dist}
\end{subfigure}
\caption{Action distribution comparison}
\label{fig:disc_dist}
\end{figure*}

In reality, whether a passenger go from ride inquiry stage to actual ride request is either yes or no, however, boolean variable is not convenient for us to model the changes in the customers' call-rate. Hence, based on historical data, we build a model to simulate the transitions and turn the boolean variable into ECR values. ECR ranges from 0 to 1 exclusive of boundaries. Before using estimated probability in our DRL algorithm training, we first evaluate the performance compared to historical data. 

In Figure \ref{fig:ecrsim}, for rides happening in each 30-minutes interval in a consecutive 14 days, we sum up the ride fares. The x axis represents each semi-hour and ranges from 1 to 48. The y axis is the sum of all the fares during the 14 days, we name it as Gross Merchandise Volume (GMV). In real world situation, the expected revenue contribution of each trip inquiry equals its ride fee times the ecr\_label, which is the boolean variable of whether a passenger turns from ride inquiry to ride request, and then times the applied discount. The results are plotted in orange and labeled as real. Meanwhile, for the same number of rides, we sum up the total simulation fare. The simulation fare of each trip is calculated as the product of the ride fee, ecr, and the applied discount of each trip. Ecr is our modeled probability of the passenger turning from ride inquiry to ride request under the applied discount of each trip. The results are plotted in blue and labeled as simulation. We can find that the simulated GMV trend is basically consistent with the real historical GMV trend. There will be slight differences in some periods, but the overall simulation error is still relatively low.

In real world application, when a passenger decide not to request a ride order after inquiring the price, the gain of the trip is zero. The same trip in our simulation, would have a gain larger than zero, because the model probability is always larger than zero. Reversely, if a passenger decide to place a request, the ecr\_label is one. In this case, the product equals the ride fee times the applied discount. Under the same situation, since our modeled probability, ecr, is less than one, the fare in simulation is smaller than the fare in reality. Ideally, when we have a very large numbers of trips, the sum of their total fares would be very close under two different modes by Law of Large Numbers. Based on Figure \ref{fig:ecrsim}, we can discover that, during the 48 semi hours, our ecr model has a close performance compared to historical real data. Our simulation values are slightly lower after 9:00am, but the differences are within a reasonable range.

In addition to ecr, we also do the same simulation on complete-rate (cr) to model the actual cr\label, which is the boolean variable indicating whether a driver has successfully answered the call and finished the ride. Similar to ecr model, in Figure \ref{fig:crsim}, our simulation values are bit lower than the real data. But the differences are acceptable when averaged over millions of rides over two weeks. 

\begin{table*}[t]
\caption{Total trip fare comparison}
\label{tripfare}
\vskip 0.15in
\begin{center}
\begin{small}
\begin{sc}
\begin{tabular}{ccccccc}
\toprule
Algorithms & Historical & IP Policy(ours) & BCQ & BEAR & BCQ+IP & BEAR+IP \\
\midrule
Budget(weekdays) & 471489.1 & 471489.07 & 320090.79 & 330953.68 & 471489.07 & 471489.06 \\

Total trip fare(weekdays) & 9935198.67 & 10304260.73 & 9839823.21 & 9850487.36 & 10302276.97 & 10238700.15 \\

Increase(weekdays) & - & 3.7\% & -0.96\%  & - 0.85\% & 3.7\% & 3.1\% \\

Budget(weekends) & 132813.58 & 132813.58 & 109123.81 & 115062.95 & 132813.57 & 132813.57 \\

Total trip fare(weekends) & 4205931.14 & 4339827.87 & 4178529.71 & 4182857.98 & 4339146.09 & 4314800.63 \\

Increase(weekends) & - & 3.18\% & -0.65\% & -0.55\% & 3.17\% & 2.59\% \\
\bottomrule
\end{tabular}
\end{sc}
\end{small}
\end{center}
\vskip -0.1in
\end{table*}

\begin{figure*}[h!tb]
\centering
\begin{subfigure}{0.4\textwidth}
	\centering
	\includegraphics[width=0.95\linewidth]{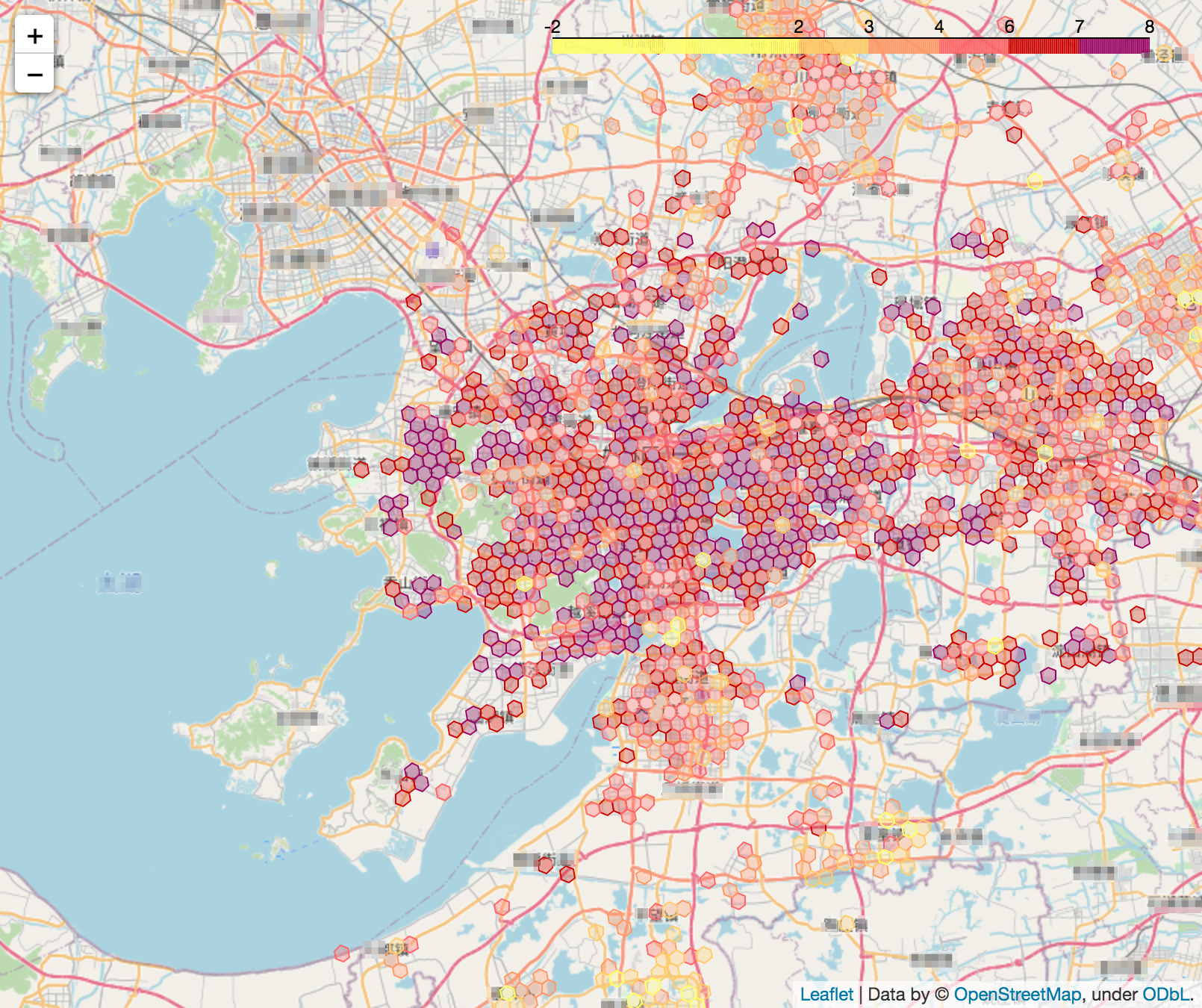}
	\caption{8:30am-9:00am, Peak time slot}
	\label{fig:peak}
\end{subfigure}
\begin{subfigure}{0.4\textwidth}
	\centering
	\includegraphics[width=0.95\linewidth]{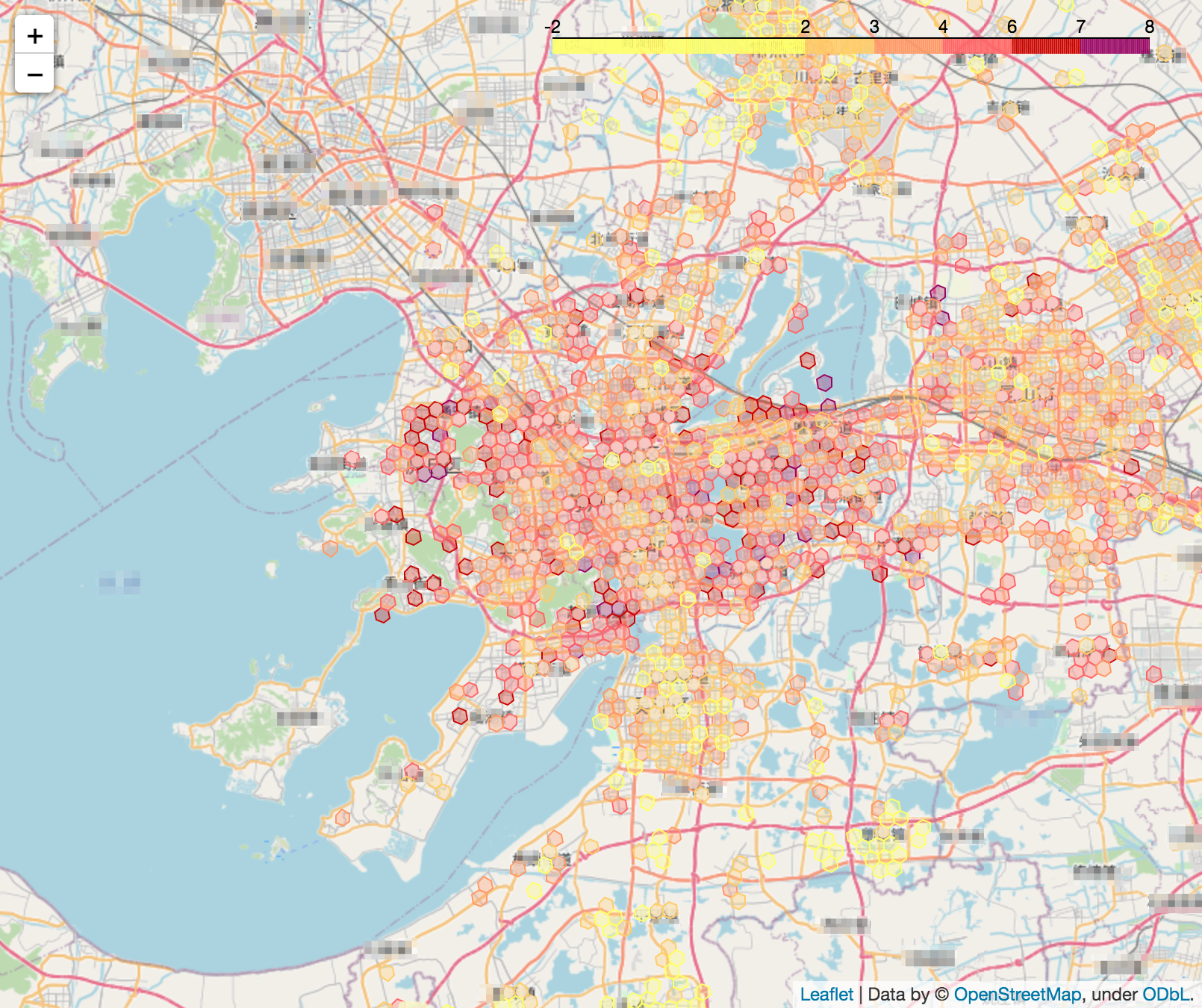}
	\caption{11:30am-12:00pm, Off-peak time slot}
    \label{fig:offpeak}
\end{subfigure}
\caption{Spatio-temporal values across the city during peak and off-peak time slots}
\label{fig:peak_vs_off_peak}
\end{figure*}

\subsection{Algorithm Performance}
We use one week of historical trip data in one city obtained from 
a major ride-hailing platform dispatching platform as our training data for learning the Spatio-Temporal network. We build the Spatio-Temporal network with two hidden dense layers and ReLU activation functions. As mentioned in Section \ref{alg:ip}, the Spatio-Temporal network $\mathcal{V}$ is then used to solve an IP policy. When solving the policy, we use $\beta = 0.5$.  

To evaluate our policy, we solve the IP policy on a week of historical unseen trip data and compare discounts assigned by our policy, with historical discounts, where historical discounts are taken according to the pricing strategy used at that time. Because our problem is under the batch DRL setting, which is purely off-line with no interactions with the environment, we pick two off-line DRL algorithms BCQ\cite{fujimoto2019offpolicy} and BEAR\cite{kumar2019stabilizing} for benchmarks. Since BCQ and BEAR do not include an Integer Programming stage to limit the budget for discounting, if we use the same reward definition as mentioned in equation \ref{eqn:rew}, we would result in a policy assigning largest discounts to all trips. Hence, when training BCQ and BEAR algorithms, we add a penalty term to each reward to indicate how much we need to compensate for distributing the discount, the penalty term is balanced by $\alpha$:
\begin{equation}
\label{eqn:rew2}
    R_{\hat{\gamma}} = R_\gamma - \alpha * (1 - a) * GMV
\end{equation}
where $a$ represents the discrete discount actions, and takes value of 0.75, 0.8, 0.85, 0.9, 0.95 or 1. 

\begin{table*}[t]
\caption{Sum of destination delta ecr values of grids in short supply}
\label{sum_dest_delta_ecr}
\vskip 0.15in
\begin{center}
\begin{small}
\begin{sc}
\begin{tabular}{ccccccc}
\toprule
Algorithms & Historical & IP Policy(ours) & BCQ & BEAR & BCQ+IP & BEAR+IP \\
\midrule
$\sum D_t^g$(weekdays) & 302.47 & 625.10 & 307.17 & 299.36 & 671.27 & 506.76 \\

Increase & - & 106.67\% & 1.55\% & -1.02\% & 122\% & 67.5\% \\

$\sum D_t^g$(weekends) & 72.49 & 182.63 & 102.35 & 102.87 & 200.75 & 139.18 \\

Increase & - & 152\% & 41.19\% &  41.90\% & 177\% & 91.99\% \\
\bottomrule
\end{tabular}
\end{sc}
\end{small}
\end{center}
\vskip -0.1in
\end{table*}

\begin{figure*}[h!tb]
\centering
\begin{subfigure}{0.3\textwidth}
	\centering
	\includegraphics[width=0.95\linewidth]{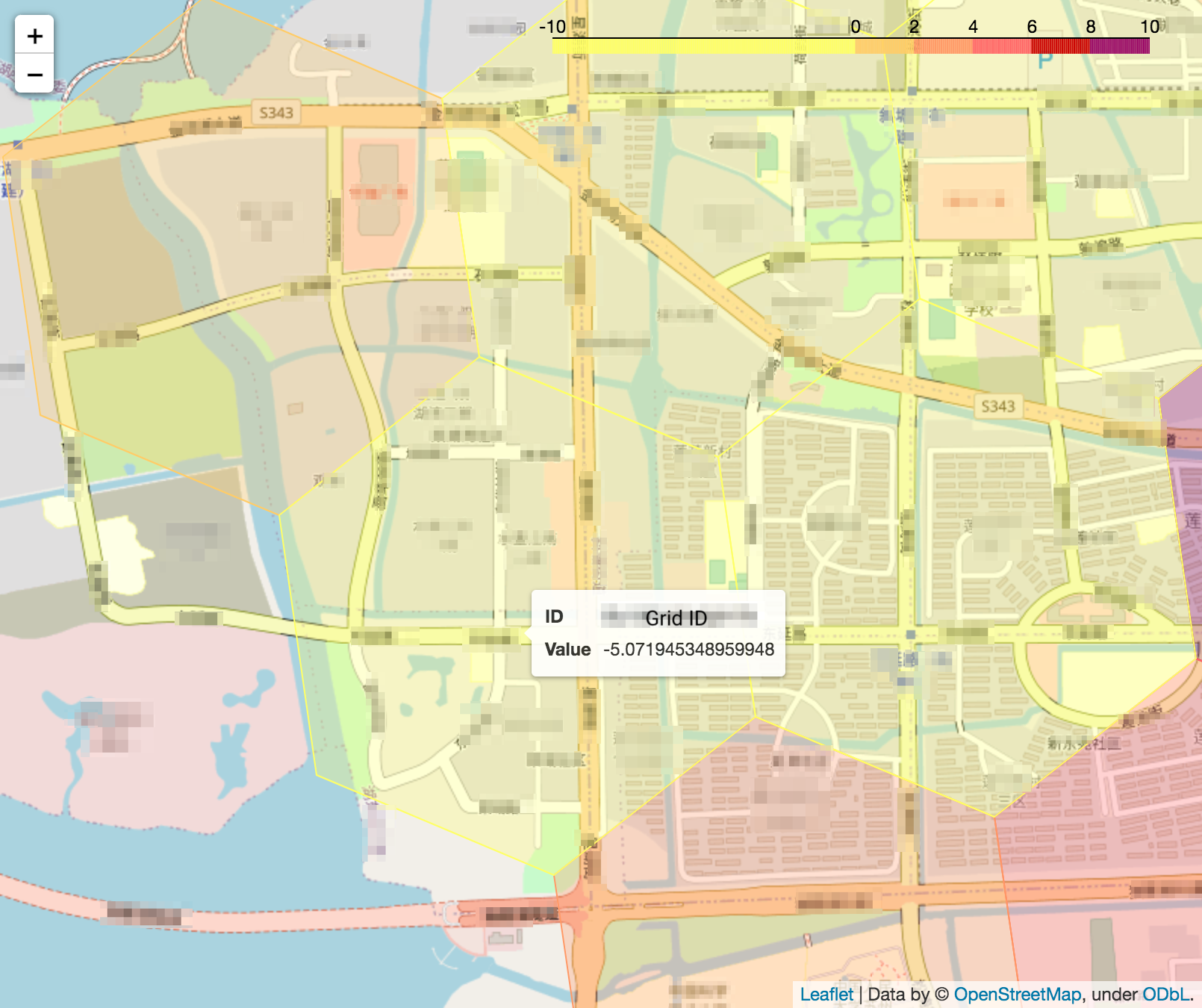}
	\caption{Value of supply minus demand}
	\label{fig:sup_dem}
\end{subfigure}
\begin{subfigure}{0.3\textwidth}
	\centering
	\includegraphics[width=0.95\linewidth]{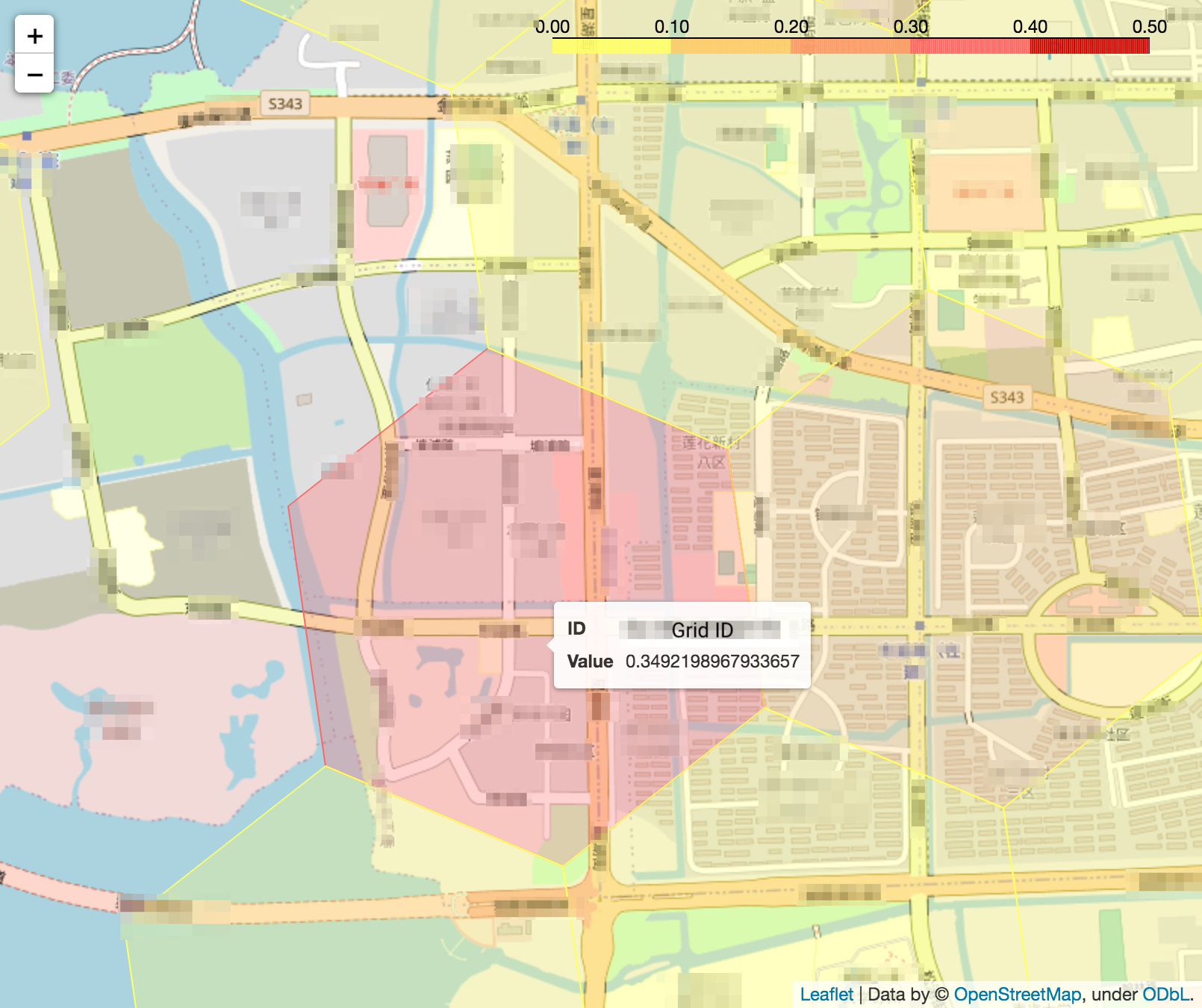}
	\caption{$D_t^g$ value given by our policy}
    \label{fig:policy}
\end{subfigure}
\begin{subfigure}{0.3\textwidth}
	\centering
	\includegraphics[width=0.95\linewidth]{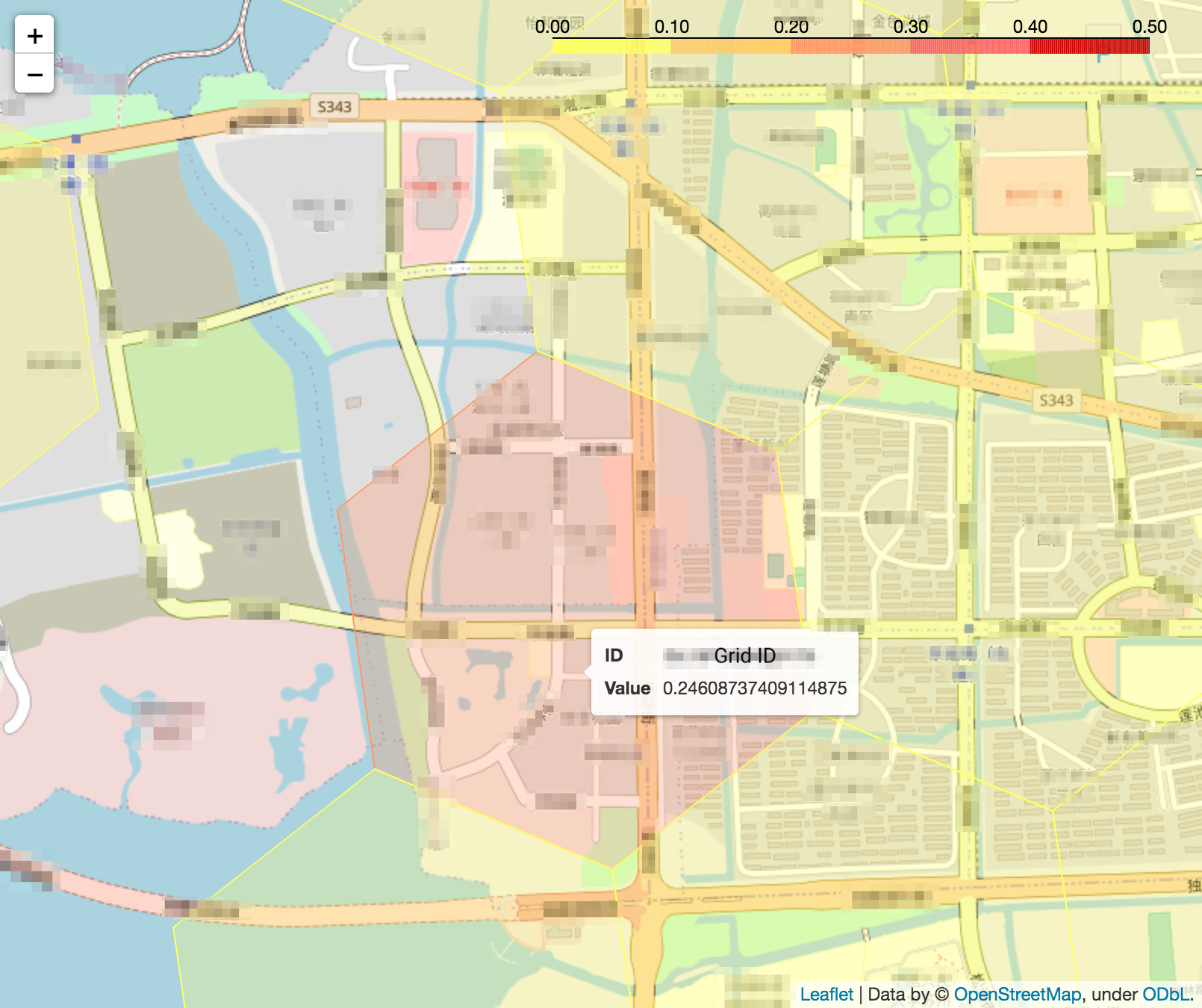}
	\caption{$D_t^g$ value given by historical data}
	\label{fig:hist}
\end{subfigure}
\caption{$D_t^g$ comparisons between our policy and historical data (Due to data security issues, we blur out the real grid IDs and names shown on the map.)}
\label{fig:dest_delta_ecr}
\end{figure*}

Figure \ref{fig:all_dist} shows the distribution over discounts by different policies trained using different algorithms. Historical data has the largest number of rides with no discounts(label = 100), 20\% off(label = 80) and 25\% off(label = 75). In other words, historical data turns to make more extreme actions to assign larger discounts. 
Our policy assigns more 5\%-off(label = 95) and 10\% off(label = 90) discounts
to the rides comparing to historical data.
On the contrary, our IP policy barely assigns 25\% off actions to the trips.
Discounts assigned by BCQ are mainly gathered at 5\%-off, while BEAR algorithm has the largest number of 10\%-off rides. To study the effect of Integer Programming(IP), we also combine BCQ and BEAR with IP, where we train BCQ and BEAR as normal, and then instead of directly using their policies during evaluation, we train an IP policy using the Q network trained by the algorithms. The results are labeled as BCQ\_IP and BEAR\_IP.
The combination has brought the discount distribution gathering around 5\% of BCQ and BEAR down to a level similar to our policy and increase the distribution on other discounts.

In addition to the overall discount distribution, we also analyze the discounts associated with trip fares. In Figure\ref{fig:policy_fare_dist} and Figure\ref{fig:hist_fare_dist}, for different trip fare intervals, we plot the discount distribution over the six discrete actions. For example, for rides with fare between 70 and 80, our policy turn to assign 20\%-off discounts to more than half of the trips. Meanwhile, historical discounts only assign about 10\% of the trips with 20\%-off discounts. Also, historical discount distribution assigns least amount of 5\% off discounts to short rides with fare between 10 and 20. On the contrary, our policy assigns least amount of 5\% off actions to long rides with fare between 70 and 80.

We summarize and compare model performance of our policy, BCQ, BCQ\_IP, BEAR and BEAR\_IP algorithms to historical data in terms of total revenue in table \ref{tripfare}. The results are evaluated over one week of unseen data. Our policy gives 3.7\% increase over the historical weekday data and 3.18\% increase over weekends. 
As we expected, other batch DRL algorithms such as BCQ and BEAR, which learn policy networks to make decisions, are basically impossible to precisely control the budget they use. In order to make sure they do not exceed the budget, we tune the penalty weight $\alpha$ in equation \ref{eqn:rew2} so as to discourage the policy from making large discount actions. Intuitively, the performance would be negatively effected. From the experiments, we can see that neither weekends nor weekday performance of BCQ or BEAR is better than historical data. To help with the decision-making process and taking better advantages of the value networks, we take the Q networks learned by BCQ and BEAR, but instead of optimizing the policy network by maximizing expected Q values, we use the Integer Programming method mentioned in section \ref{sec:alg} as their behavior policy. The results are labeled as BCQ\_IP and BEAR\_IP respectively. By taking this modification, we witness improvements over weekdays from -0.96\% to 3.7\% for BCQ and -0.85\% to 3.1\% for BEAR and over weekends from -0.65\% to 3.17\% for BCQ and -0.55\% to 2.59\% for BEAR.

\subsection{Visualization on City Maps}
In addition to quantitative analysis and results, we also discover interesting phenomenons by visualizing our spatio-temporal value net on the city map. For example, Figure~\ref{fig:peak_vs_off_peak} shows the spatio temporal values of each grid across the city, at two different time slots. Figure \ref{fig:peak} shows the spatio-temporal value distribution at peak time from 8:30am to 9:00am. Figure\ref{fig:offpeak} shows the spatio-temporal values of the same city but at the off-peak hours from 11:30am to 12:00pm. Not surprisingly, during the morning peak time slot, the whole city is more busy and more rides are happening compared to the off-peak time slot. Downtown areas are especially more valuable than other places in the city.

To see how our algorithm helps to balance the supply and demand situation of the city. We introduce a new quantitative feature called Dest\_delta\_ecr, and denote it as $D$. The feature is calculated for each grid of the city at each time slot t. It is defined as $ D_t^g = \sum_{i=1}^{M_{t}} \delta \text{ecr}_i$, where $M_t$ is the set of all the rides arriving at destination grid $g$ at time slot $t$ and $\delta \text{ecr}_i$ is the change in $\text{ecr}$ of each ride $i$ in the set $M_t$ under its assigned discount.

We also calculate supply minus demand information of each hexagon grid of the city map at each time slot t. If supply minus demand is smaller than zero, we can say that at this time slot t, the grid is in short supply and needs more drivers. For example, in Figure \ref{fig:sup_dem}, the grid which we are focused on is a residential area. During the morning peak time slot on a weekday, lots of people are leaving the area for work, thus, the grid has a negative value of supply minus demand, which means it's in short supply. Hence, we want to encourage rides to the area at that time slot, which will also bring drivers to the area to compensate for the short supply.  

Under this situation, historical data, as shown in figure \ref{fig:hist} has $D_t$ equals 0.246, while our policy in Figure \ref{fig:policy} has $D_t$ equals 0.383 at the same grid and same time slot. This means, discount distribution by our policy, compared to historical distribution, assigns more discounts on trips going to that residential grid. In other words, we increase the passengers' ecr of the rides going to the grid at that time slot in order to help with the supply and demand imbalance of the city. 

To better analyze and has a broader view of the effectiveness of our policy on the imbalanced supply and demand problem, we sum over all $D_t^g$ only on destination grids that are in short supply. Table \ref{sum_dest_delta_ecr} summarizes the results. We want the grids in short supply to have more drivers coming to meet the demands, which means we expect the sum over all $D_t^g$ for grids in short supply to be as large as possible but meanwhile do not exceed the budget. 
For weekdays, our policy has increased the sum of destination delta ecr values by 106.67\%, and has increased the sum of destination delta ecr values by 152\% for weekends. BCQ and BEAR have also increased about 40\% for weekends, but BCQ only does a slightly better job for weekdays. However, when combining with IP, BCQ\_IP boosts the performance by 122\% for weekdays and 177\% for weekends. BEAR\_IP also does a much better job than vanilla BEAR. The result indicates the importance of IP and the advantage over a normal policy network in our application of discount distribution.

\section{Conclusion}
\label{sec:con}
This paper has proposed an offline deep reinforcement learning method to learn the spatial-temporal long-term value in the whole city, as a guidance to dynamically adjust the spatial-temporal distribution of demand, to improve the efficiency of the marketplace. The goal is to optimize the ride-hailing platform's long-term efficiency, as well as balancing customer demands and driver supply in the city. To do so, we model the pricing problem as a sequential decision-making problem. Experiments demonstrate the effectiveness of the proposed algorithm and the importance of the integer programming step. We deliver an improvements in both total revenue and efficiency in balancing supply and demand across the city. 






\bibliographystyle{ACM-Reference-Format}
\bibliography{didi}


\begin{thebibliography}{26}


\ifx \showCODEN    \undefined \def \showCODEN     #1{\unskip}     \fi
\ifx \showDOI      \undefined \def \showDOI       #1{#1}\fi
\ifx \showISBNx    \undefined \def \showISBNx     #1{\unskip}     \fi
\ifx \showISBNxiii \undefined \def \showISBNxiii  #1{\unskip}     \fi
\ifx \showISSN     \undefined \def \showISSN      #1{\unskip}     \fi
\ifx \showLCCN     \undefined \def \showLCCN      #1{\unskip}     \fi
\ifx \shownote     \undefined \def \shownote      #1{#1}          \fi
\ifx \showarticletitle \undefined \def \showarticletitle #1{#1}   \fi
\ifx \showURL      \undefined \def \showURL       {\relax}        \fi
\providecommand\bibfield[2]{#2}
\providecommand\bibinfo[2]{#2}
\providecommand\natexlab[1]{#1}
\providecommand\showeprint[2][]{arXiv:#2}

\bibitem[\protect\citeauthoryear{Bao, He, Ruan, Li, and Zheng}{Bao
  et~al\mbox{.}}{2017}]%
        {bikelane}
\bibfield{author}{\bibinfo{person}{Jie Bao}, \bibinfo{person}{Tianfu He},
  \bibinfo{person}{Sijie Ruan}, \bibinfo{person}{Yanhua Li}, {and}
  \bibinfo{person}{Yu Zheng}.} \bibinfo{year}{2017}\natexlab{}.
\newblock \showarticletitle{Planning Bike Lanes Based on Sharing-Bikes'
  Trajectories}. In \bibinfo{booktitle}{\emph{Proceedings of the 23rd ACM
  SIGKDD International Conference on Knowledge Discovery and Data Mining}}
  (Halifax, NS, Canada) \emph{(\bibinfo{series}{KDD '17})}.
  \bibinfo{address}{New York, NY, USA}, \bibinfo{pages}{1377–1386}.
\newblock


\bibitem[\protect\citeauthoryear{Bimpikis, Candogan, and Saban}{Bimpikis
  et~al\mbox{.}}{2017}]%
        {Bimpikis2017}
\bibfield{author}{\bibinfo{person}{Kostas Bimpikis}, \bibinfo{person}{Ozan
  Candogan}, {and} \bibinfo{person}{Daniela Saban}.}
  \bibinfo{year}{2017}\natexlab{}.
\newblock \showarticletitle{Spatial Pricing in Ride-Sharing Networks}. In
  \bibinfo{booktitle}{\emph{Proceedings of the 12th Workshop on the Economics
  of Networks, Systems and Computation}} \emph{(\bibinfo{series}{NetEcon
  '17})}. \bibinfo{publisher}{Association for Computing Machinery},
  \bibinfo{address}{New York, NY, USA}, Article \bibinfo{articleno}{5},
  \bibinfo{numpages}{1}~pages.
\newblock
\showISBNx{9781450350891}


\bibitem[\protect\citeauthoryear{Chen, Yao, Mo, Zhu, and Chen}{Chen
  et~al\mbox{.}}{2021}]%
        {CHEN2021103272}
\bibfield{author}{\bibinfo{person}{Chuqiao Chen}, \bibinfo{person}{Fugen Yao},
  \bibinfo{person}{Dong Mo}, \bibinfo{person}{Jiangtao Zhu}, {and}
  \bibinfo{person}{Xiqun~(Michael) Chen}.} \bibinfo{year}{2021}\natexlab{}.
\newblock \showarticletitle{Spatial-temporal pricing for ride-sourcing platform
  with reinforcement learning}.
\newblock \bibinfo{journal}{\emph{Transportation Research Part C: Emerging
  Technologies}}  \bibinfo{volume}{130} (\bibinfo{year}{2021}),
  \bibinfo{pages}{103272}.
\newblock
\showISSN{0968-090X}


\bibitem[\protect\citeauthoryear{Fujimoto, Meger, and Precup}{Fujimoto
  et~al\mbox{.}}{2019a}]%
        {fujimoto2019off}
\bibfield{author}{\bibinfo{person}{Scott Fujimoto}, \bibinfo{person}{David
  Meger}, {and} \bibinfo{person}{Doina Precup}.}
  \bibinfo{year}{2019}\natexlab{a}.
\newblock \showarticletitle{Off-policy deep reinforcement learning without
  exploration}. In \bibinfo{booktitle}{\emph{International Conference on
  Machine Learning}}. PMLR, \bibinfo{pages}{2052--2062}.
\newblock


\bibitem[\protect\citeauthoryear{Fujimoto, Meger, and Precup}{Fujimoto
  et~al\mbox{.}}{2019b}]%
        {fujimoto2019offpolicy}
\bibfield{author}{\bibinfo{person}{Scott Fujimoto}, \bibinfo{person}{David
  Meger}, {and} \bibinfo{person}{Doina Precup}.}
  \bibinfo{year}{2019}\natexlab{b}.
\newblock \showarticletitle{Off-policy deep reinforcement learning without
  exploration}. In \bibinfo{booktitle}{\emph{International Conference on
  Machine Learning}}. PMLR, \bibinfo{pages}{2052--2062}.
\newblock


\bibitem[\protect\citeauthoryear{Glaschenko, Ivaschenko, Rzevski, and
  Skobelev}{Glaschenko et~al\mbox{.}}{2009}]%
        {Glaschenko2009MultiAgentRT}
\bibfield{author}{\bibinfo{person}{Andrey Glaschenko}, \bibinfo{person}{A.
  Ivaschenko}, \bibinfo{person}{G. Rzevski}, {and} \bibinfo{person}{P.
  Skobelev}.} \bibinfo{year}{2009}\natexlab{}.
\newblock \showarticletitle{Multi-Agent Real Time Scheduling System for Taxi
  Companies}.
\newblock


\bibitem[\protect\citeauthoryear{He and Shen}{He and Shen}{2015}]%
        {HE201593}
\bibfield{author}{\bibinfo{person}{Fang He} {and} \bibinfo{person}{Zuo-Jun~Max
  Shen}.} \bibinfo{year}{2015}\natexlab{}.
\newblock \showarticletitle{Modeling taxi services with smartphone-based
  e-hailing applications}.
\newblock \bibinfo{journal}{\emph{Transportation Research Part C: Emerging
  Technologies}}  \bibinfo{volume}{58} (\bibinfo{year}{2015}),
  \bibinfo{pages}{93--106}.
\newblock
\showISSN{0968-090X}


\bibitem[\protect\citeauthoryear{Kumar, Fu, Soh, Tucker, and Levine}{Kumar
  et~al\mbox{.}}{2019}]%
        {kumar2019stabilizing}
\bibfield{author}{\bibinfo{person}{Aviral Kumar}, \bibinfo{person}{Justin Fu},
  \bibinfo{person}{Matthew Soh}, \bibinfo{person}{George Tucker}, {and}
  \bibinfo{person}{Sergey Levine}.} \bibinfo{year}{2019}\natexlab{}.
\newblock \showarticletitle{Stabilizing Off-Policy Q-Learning via Bootstrapping
  Error Reduction}.
\newblock \bibinfo{journal}{\emph{Advances in Neural Information Processing
  Systems}}  \bibinfo{volume}{32} (\bibinfo{year}{2019}),
  \bibinfo{pages}{11784--11794}.
\newblock


\bibitem[\protect\citeauthoryear{Kumar, Zhou, Tucker, and Levine}{Kumar
  et~al\mbox{.}}{2020}]%
        {kumar2020conservative}
\bibfield{author}{\bibinfo{person}{Aviral Kumar}, \bibinfo{person}{Aurick
  Zhou}, \bibinfo{person}{George Tucker}, {and} \bibinfo{person}{Sergey
  Levine}.} \bibinfo{year}{2020}\natexlab{}.
\newblock \showarticletitle{Conservative q-learning for offline reinforcement
  learning}.
\newblock \bibinfo{journal}{\emph{arXiv preprint arXiv:2006.04779}}
  (\bibinfo{year}{2020}).
\newblock


\bibitem[\protect\citeauthoryear{Levine, Kumar, Tucker, and Fu}{Levine
  et~al\mbox{.}}{2020}]%
        {levine2020offline}
\bibfield{author}{\bibinfo{person}{Sergey Levine}, \bibinfo{person}{Aviral
  Kumar}, \bibinfo{person}{George Tucker}, {and} \bibinfo{person}{Justin Fu}.}
  \bibinfo{year}{2020}\natexlab{}.
\newblock \showarticletitle{Offline reinforcement learning: Tutorial, review,
  and perspectives on open problems}.
\newblock \bibinfo{journal}{\emph{arXiv preprint arXiv:2005.01643}}
  (\bibinfo{year}{2020}).
\newblock


\bibitem[\protect\citeauthoryear{Li, Qin, Jiao, Yang, Wang, Wang, Wu, and
  Ye}{Li et~al\mbox{.}}{2019}]%
        {li2019efficient}
\bibfield{author}{\bibinfo{person}{Minne Li}, \bibinfo{person}{Zhiwei Qin},
  \bibinfo{person}{Yan Jiao}, \bibinfo{person}{Yaodong Yang},
  \bibinfo{person}{Jun Wang}, \bibinfo{person}{Chenxi Wang},
  \bibinfo{person}{Guobin Wu}, {and} \bibinfo{person}{Jieping Ye}.}
  \bibinfo{year}{2019}\natexlab{}.
\newblock \showarticletitle{Efficient ridesharing order dispatching with mean
  field multi-agent reinforcement learning}. In \bibinfo{booktitle}{\emph{The
  World Wide Web Conference}}. \bibinfo{pages}{983--994}.
\newblock


\bibitem[\protect\citeauthoryear{Li, Zeng, Yang, and Zhang}{Li
  et~al\mbox{.}}{2009}]%
        {route1}
\bibfield{author}{\bibinfo{person}{Qingquan Li}, \bibinfo{person}{Zhe Zeng},
  \bibinfo{person}{Bisheng Yang}, {and} \bibinfo{person}{Tong Zhang}.}
  \bibinfo{year}{2009}\natexlab{}.
\newblock \showarticletitle{Hierarchical route planning based on taxi
  GPS-trajectories}. In \bibinfo{booktitle}{\emph{2009 17th International
  Conference on Geoinformatics}}. \bibinfo{pages}{1--5}.
\newblock


\bibitem[\protect\citeauthoryear{Mnih, Kavukcuoglu, Silver, Rusu, Veness,
  Bellemare, Graves, Riedmiller, Fidjeland, Ostrovski, Petersen, Beattie,
  Sadik, Antonoglou, King, Kumaran, Wierstra, Legg, and Hassabis}{Mnih
  et~al\mbox{.}}{2015}]%
        {mnih2015humanlevel}
\bibfield{author}{\bibinfo{person}{Volodymyr Mnih}, \bibinfo{person}{Koray
  Kavukcuoglu}, \bibinfo{person}{David Silver}, \bibinfo{person}{Andrei~A.
  Rusu}, \bibinfo{person}{Joel Veness}, \bibinfo{person}{Marc~G. Bellemare},
  \bibinfo{person}{Alex Graves}, \bibinfo{person}{Martin Riedmiller},
  \bibinfo{person}{Andreas~K. Fidjeland}, \bibinfo{person}{Georg Ostrovski},
  \bibinfo{person}{Stig Petersen}, \bibinfo{person}{Charles Beattie},
  \bibinfo{person}{Amir Sadik}, \bibinfo{person}{Ioannis Antonoglou},
  \bibinfo{person}{Helen King}, \bibinfo{person}{Dharshan Kumaran},
  \bibinfo{person}{Daan Wierstra}, \bibinfo{person}{Shane Legg}, {and}
  \bibinfo{person}{Demis Hassabis}.} \bibinfo{year}{2015}\natexlab{}.
\newblock \showarticletitle{Human-level control through deep reinforcement
  learning}.
\newblock \bibinfo{journal}{\emph{Nature}} \bibinfo{volume}{518},
  \bibinfo{number}{7540} (\bibinfo{date}{Feb.} \bibinfo{year}{2015}),
  \bibinfo{pages}{529--533}.
\newblock
\showISSN{00280836}


\bibitem[\protect\citeauthoryear{Nourinejad and Ramezani}{Nourinejad and
  Ramezani}{2020}]%
        {NOURINEJAD2020340}
\bibfield{author}{\bibinfo{person}{Mehdi Nourinejad} {and}
  \bibinfo{person}{Mohsen Ramezani}.} \bibinfo{year}{2020}\natexlab{}.
\newblock \showarticletitle{Ride-Sourcing modeling and pricing in
  non-equilibrium two-sided markets}.
\newblock \bibinfo{journal}{\emph{Transportation Research Part B:
  Methodological}}  \bibinfo{volume}{132} (\bibinfo{year}{2020}),
  \bibinfo{pages}{340--357}.
\newblock
\showISSN{0191-2615}
\urldef\tempurl%
\url{https://doi.org/10.1016/j.trb.2019.05.019}
\showDOI{\tempurl}
\newblock
\shownote{23rd International Symposium on Transportation and Traffic Theory
  (ISTTT 23)}.


\bibitem[\protect\citeauthoryear{Shang, Li, Qin, Yu, Meng, and Ye}{Shang
  et~al\mbox{.}}{2021}]%
        {shang2021partially}
\bibfield{author}{\bibinfo{person}{Wenjie Shang}, \bibinfo{person}{Qingyang
  Li}, \bibinfo{person}{Zhiwei Qin}, \bibinfo{person}{Yang Yu},
  \bibinfo{person}{Yiping Meng}, {and} \bibinfo{person}{Jieping Ye}.}
  \bibinfo{year}{2021}\natexlab{}.
\newblock \showarticletitle{Partially observable environment estimation with
  uplift inference for reinforcement learning based recommendation}.
\newblock \bibinfo{journal}{\emph{Machine Learning}} (\bibinfo{year}{2021}),
  \bibinfo{pages}{1--38}.
\newblock


\bibitem[\protect\citeauthoryear{Shang, Yu, Li, Qin, Meng, and Ye}{Shang
  et~al\mbox{.}}{2019}]%
        {shang2019environment}
\bibfield{author}{\bibinfo{person}{Wenjie Shang}, \bibinfo{person}{Yang Yu},
  \bibinfo{person}{Qingyang Li}, \bibinfo{person}{Zhiwei Qin},
  \bibinfo{person}{Yiping Meng}, {and} \bibinfo{person}{Jieping Ye}.}
  \bibinfo{year}{2019}\natexlab{}.
\newblock \showarticletitle{Environment reconstruction with hidden confounders
  for reinforcement learning based recommendation}. In
  \bibinfo{booktitle}{\emph{Proceedings of the 25th ACM SIGKDD International
  Conference on Knowledge Discovery \& Data Mining}}.
  \bibinfo{pages}{566--576}.
\newblock


\bibitem[\protect\citeauthoryear{Tang, Qin, Zhang, Wang, Xu, Ma, Zhu, and
  Ye}{Tang et~al\mbox{.}}{2021}]%
        {tang2021deep}
\bibfield{author}{\bibinfo{person}{Xiaocheng Tang}, \bibinfo{person}{Zhiwei
  Qin}, \bibinfo{person}{Fan Zhang}, \bibinfo{person}{Zhaodong Wang},
  \bibinfo{person}{Zhe Xu}, \bibinfo{person}{Yintai Ma},
  \bibinfo{person}{Hongtu Zhu}, {and} \bibinfo{person}{Jieping Ye}.}
  \bibinfo{year}{2021}\natexlab{}.
\newblock \bibinfo{title}{A Deep Value-network Based Approach for Multi-Driver
  Order Dispatching}.
\newblock
\newblock
\showeprint[arxiv]{2106.04493}~[cs.LG]


\bibitem[\protect\citeauthoryear{Tong, Chen, Zhou, Chen, Wang, Yang, Ye, and
  Lv}{Tong et~al\mbox{.}}{2017}]%
        {supplychain}
\bibfield{author}{\bibinfo{person}{Yongxin Tong}, \bibinfo{person}{Yuqiang
  Chen}, \bibinfo{person}{Zimu Zhou}, \bibinfo{person}{Lei Chen},
  \bibinfo{person}{Jie Wang}, \bibinfo{person}{Qiang Yang},
  \bibinfo{person}{Jieping Ye}, {and} \bibinfo{person}{Weifeng Lv}.}
  \bibinfo{year}{2017}\natexlab{}.
\newblock \showarticletitle{The Simpler The Better: A Unified Approach to
  Predicting Original Taxi Demands Based on Large-Scale Online Platforms}. In
  \bibinfo{booktitle}{\emph{Proceedings of the 23rd ACM SIGKDD International
  Conference on Knowledge Discovery and Data Mining}} (Halifax, NS, Canada)
  \emph{(\bibinfo{series}{KDD '17})}. \bibinfo{address}{New York, NY, USA},
  \bibinfo{pages}{1653–1662}.
\newblock


\bibitem[\protect\citeauthoryear{van Hasselt, Guez, and Silver}{van Hasselt
  et~al\mbox{.}}{2015}]%
        {vanhasselt2015deep}
\bibfield{author}{\bibinfo{person}{Hado van Hasselt}, \bibinfo{person}{Arthur
  Guez}, {and} \bibinfo{person}{David Silver}.}
  \bibinfo{year}{2015}\natexlab{}.
\newblock \bibinfo{title}{Deep Reinforcement Learning with Double Q-learning}.
\newblock
\newblock
\showeprint[arxiv]{1509.06461}~[cs.LG]


\bibitem[\protect\citeauthoryear{Wang, Qin, Tang, Ye, and Zhu}{Wang
  et~al\mbox{.}}{2018}]%
        {drlonline}
\bibfield{author}{\bibinfo{person}{Zhaodong Wang}, \bibinfo{person}{Zhiwei
  Qin}, \bibinfo{person}{Xiaocheng Tang}, \bibinfo{person}{Jieping Ye}, {and}
  \bibinfo{person}{Hongtu Zhu}.} \bibinfo{year}{2018}\natexlab{}.
\newblock \showarticletitle{Deep Reinforcement Learning with Knowledge Transfer
  for Online Rides Order Dispatching}. In \bibinfo{booktitle}{\emph{2018 IEEE
  International Conference on Data Mining (ICDM)}}. \bibinfo{pages}{617--626}.
\newblock


\bibitem[\protect\citeauthoryear{Wu, Srikant, Liu, and Jiang}{Wu
  et~al\mbox{.}}{2015}]%
        {wu2015algorithms}
\bibfield{author}{\bibinfo{person}{Huasen Wu}, \bibinfo{person}{R Srikant},
  \bibinfo{person}{Xin Liu}, {and} \bibinfo{person}{Chong Jiang}.}
  \bibinfo{year}{2015}\natexlab{}.
\newblock \showarticletitle{Algorithms with Logarithmic or Sublinear Regret for
  Constrained Contextual Bandits}.
\newblock \bibinfo{journal}{\emph{Advances in Neural Information Processing
  Systems}}  \bibinfo{volume}{28} (\bibinfo{year}{2015}),
  \bibinfo{pages}{433--441}.
\newblock


\bibitem[\protect\citeauthoryear{Xin-min, Yu-ting, and Song-chen}{Xin-min
  et~al\mbox{.}}{2010}]%
        {route2}
\bibfield{author}{\bibinfo{person}{Tang Xin-min}, \bibinfo{person}{Wang
  Yu-ting}, {and} \bibinfo{person}{Han Song-chen}.}
  \bibinfo{year}{2010}\natexlab{}.
\newblock \showarticletitle{Aircraft Taxi Route Planning for A-SMGCS Based on
  Discrete Event Dynamic System modeling}. In \bibinfo{booktitle}{\emph{2010
  Second International Conference on Computer Modeling and Simulation}},
  Vol.~\bibinfo{volume}{1}. \bibinfo{pages}{224--228}.
\newblock


\bibitem[\protect\citeauthoryear{Xu, Li, Guan, Zhang, Li, Nan, Liu, Bian, and
  Ye}{Xu et~al\mbox{.}}{2018}]%
        {orderdispatch}
\bibfield{author}{\bibinfo{person}{Zhe Xu}, \bibinfo{person}{Zhixin Li},
  \bibinfo{person}{Qingwen Guan}, \bibinfo{person}{Dingshui Zhang},
  \bibinfo{person}{Qiang Li}, \bibinfo{person}{Junxiao Nan},
  \bibinfo{person}{Chunyang Liu}, \bibinfo{person}{Wei Bian}, {and}
  \bibinfo{person}{Jieping Ye}.} \bibinfo{year}{2018}\natexlab{}.
\newblock \showarticletitle{Large-Scale Order Dispatch in On-Demand
  Ride-Hailing Platforms: A Learning and Planning Approach}. In
  \bibinfo{booktitle}{\emph{Proceedings of the 24th ACM SIGKDD International
  Conference on Knowledge Discovery \& Data Mining}} (London, United Kingdom)
  \emph{(\bibinfo{series}{KDD '18})}. \bibinfo{address}{New York, NY, USA},
  \bibinfo{pages}{905–913}.
\newblock


\bibitem[\protect\citeauthoryear{Yang, Shao, Wang, and Ye}{Yang
  et~al\mbox{.}}{2020}]%
        {YANG2020126}
\bibfield{author}{\bibinfo{person}{Hai Yang}, \bibinfo{person}{Chaoyi Shao},
  \bibinfo{person}{Hai Wang}, {and} \bibinfo{person}{Jieping Ye}.}
  \bibinfo{year}{2020}\natexlab{}.
\newblock \showarticletitle{Integrated reward scheme and surge pricing in a
  ridesourcing market}.
\newblock \bibinfo{journal}{\emph{Transportation Research Part B:
  Methodological}}  \bibinfo{volume}{134} (\bibinfo{year}{2020}),
  \bibinfo{pages}{126--142}.
\newblock
\showISSN{0191-2615}


\bibitem[\protect\citeauthoryear{Yao, Wu, Ke, Tang, Jia, Lu, Gong, Ye, and
  Li}{Yao et~al\mbox{.}}{2018}]%
        {yao2018deep}
\bibfield{author}{\bibinfo{person}{Huaxiu Yao}, \bibinfo{person}{Fei Wu},
  \bibinfo{person}{Jintao Ke}, \bibinfo{person}{Xianfeng Tang},
  \bibinfo{person}{Yitian Jia}, \bibinfo{person}{Siyu Lu},
  \bibinfo{person}{Pinghua Gong}, \bibinfo{person}{Jieping Ye}, {and}
  \bibinfo{person}{Zhenhui Li}.} \bibinfo{year}{2018}\natexlab{}.
\newblock \showarticletitle{Deep multi-view spatial-temporal network for taxi
  demand prediction}. In \bibinfo{booktitle}{\emph{Proceedings of the AAAI
  Conference on Artificial Intelligence}}, Vol.~\bibinfo{volume}{32}.
\newblock


\bibitem[\protect\citeauthoryear{Zha, Yin, and Xu}{Zha et~al\mbox{.}}{2018}]%
        {ZHA201858}
\bibfield{author}{\bibinfo{person}{Liteng Zha}, \bibinfo{person}{Yafeng Yin},
  {and} \bibinfo{person}{Zhengtian Xu}.} \bibinfo{year}{2018}\natexlab{}.
\newblock \showarticletitle{Geometric matching and spatial pricing in
  ride-sourcing markets}.
\newblock \bibinfo{journal}{\emph{Transportation Research Part C: Emerging
  Technologies}}  \bibinfo{volume}{92} (\bibinfo{year}{2018}),
  \bibinfo{pages}{58--75}.
\newblock
\showISSN{0968-090X}


\end{thebibliography}

\end{document}